\documentclass[journal]{IEEEtran}
\usepackage{cite}

\ifCLASSINFOpdf
  \usepackage[pdftex]{graphicx}
  \graphicspath{{../pdf/}{../jpeg/}}
  \DeclareGraphicsExtensions{.pdf,.jpeg,.png}
\else
\fi
\usepackage{amsmath}
\usepackage{array}

 \usepackage[caption=false,font=footnotesize]{subfig}
\usepackage{url}

\usepackage{amsmath}
\usepackage{amsthm}
\usepackage{booktabs}
\usepackage{multirow}
\usepackage{amssymb}
\usepackage{algorithm}
\usepackage{algorithmic}

\newcommand{\ie}{\textit{i}.\textit{e}.}
\newcommand{\eg}{\textit{e}.\textit{g}.}

\begin{document}
%
\title{X-view: Non-egocentric Multi-View 3D Object Detector}

\author{Liang~Xie,
        Guodong~Xu,
        Deng~Cai,~\IEEEmembership{Member,~IEEE},
        Xiaofei~He,~\IEEEmembership{Senior Member,~IEEE}}
%
%
%

%
%

\markboth{Journal of \LaTeX\ Class Files}%
{Shell \MakeLowercase{\textit{et al.}}: Bare Demo of IEEEtran.cls for IEEE Journals}
%



\maketitle

\begin{abstract}
3D object detection algorithms for autonomous driving reason about 3D obstacles either from 3D birds-eye view or perspective view or both. Recent works attempt to improve the detection performance via mining and fusing from multiple egocentric views. Although the egocentric perspective view alleviates some weaknesses of the birds-eye view, the sectored grid partition becomes so coarse in the distance that the targets and surrounding context mix together, which makes the features less discriminative. In this paper, we generalize the research on 3D multi-view learning and propose a novel multi-view-based 3D detection method, named X-view, to overcome the drawbacks of the multi-view methods. Specifically, X-view breaks through the traditional limitation about the perspective view whose original point must be consistent with the 3D Cartesian coordinate. X-view is designed as a general paradigm that can be applied on almost any 3D detectors based on LiDAR with only little increment of running time, no matter it is voxel/grid-based or raw-point-based. We conduct experiments on KITTI\cite{geiger2015kitti} and NuScenes\cite{caesar2020nuscenes} datasets to demonstrate the robustness and effectiveness of our proposed X-view. The results show that X-view obtains consistent improvements when combined with four mainstream state-of-the-art 3D methods: SECOND\cite{yan2018second}, PointRCNN\cite{shi2019pointrcnn}, Part-$A^2$\cite{shi2020points}, and PV-RCNN\cite{shi2020pv}.
\end{abstract}


%
\IEEEpeerreviewmaketitle

\section{Introduction}
%
%
%
%

\IEEEPARstart{3}{D} object detection is an essential component of autonomous driving. 3D detectors identify the road obstacles to help the driving system make correct decisions to ensure safety and driving effectiveness effectively. With the rapid development and the decrease of the 3D sensor's production costs, LiDAR becomes a necessary module on the self-driving car for perceiving the scenes as it can scan the surrounding environments and capture an accurate 3D description. Unlike 2D image, point clouds generated by LiDAR have precise depth values and 3D space information, making it more reliable to reason about accurate 3D objects.

\begin{figure}[t]
	\centering
	\includegraphics[width=0.48\textwidth]{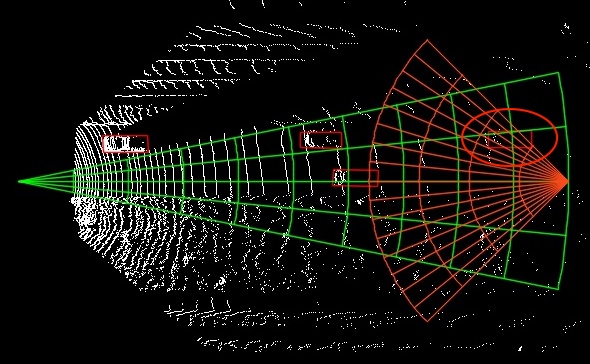}
	\caption{This figure illustrates the advantages of the non-egocentric PV(perspective view). In the figure, the green grids are the voxels of egocentric PV, and the orange grids represent a non-egocentric PV whose origin is placed in the distance. In the PV coordinate(\eg spherical or cylindrical coordinate), more closer to the origin, the grids are more meticulous, vice versa. Therefore, for the objects in the distance, \eg the object in the red circle, they are only voxelized into few grids in the egocentric PV coordinate, which brings difficulty for precisely recognizing and locating. The non-egocentric PV can remedy this issue because these objects are voxelized more meticulously in a non-egocentric view whose origin is near the objects.}
	\label{fig:non-egocentric}
\end{figure}

For the LiDAR-based methods, whatever grid-based\cite{zhou2018voxelnet, lang2019pointpillars, yan2018second} or point-based\cite{shi2019pointrcnn, shi2020points, shi2020point, wang2019dynamic}, they utilize the point clouds in the original LiDAR \textit{xyz}-coordinate whose origin is located in the LiDAR. We denote this 3D \textit{xyz}-coordinate as birds-eye view(BEV). Points in BEV capture precise 3D structures of obstacles and environments, and the shapes of objects are distance-invariant. However, for the objects in the distance or small-sized objects, points become very sparse, making detectors very hard to tackle those challenging targets. Although LiDAR points are sparse in the 3D \textit{xyz}-coordinate, they are inherently dense from the view of the sensor, \ie the density of points is consistent in angle in the perspective view\cite{meyer2019lasernet}. Another issue that the birds-eye view encounters is the local feature mismatching. The point densities vary with the distance, and it is unfriendly to be resolved by local parameter-sharing feature extractors, such as 2D/3D CNN or the architecture of PointNet++\cite{qi2017pointnet++}. This issue is remedied in the perspective view where the point densities keep consistent in angle.

Therefore, some works\cite{chen2017multi, meyer2019lasernet, zhou2020end, wang2020pillar} attempt to employ the perspective view to remedy the drawbacks of the birds-eye view. However, the methods using perspective-image\cite{chen2017multi,meyer2019lasernet} suffer the same troubles with monocular methods that the object shapes are not distant-invariant. And due to the loss of depth information, objects will overlap in 2D images, making it hard to be accurately recognized and located. Recent multi-view based approaches\cite{zhou2020end, wang2020pillar} projects points into a 3D spherical or cylindrical coordinate to fix the problem of losing depth information, and the distant-variant issue is also resolved by combining multi-view features. However, they only consider egocentric views. In the perspective view(PV), the space is voxelized into series of sectored grids along \textit{x} and \textit{y} axis and become larger and sparser as the distance increase. In the egocentric PV, the features of distant objects are mixed with the surrounding context and lose discriminability. Hence, the egocentric PV is still insufficient to tackle this issue. The solution to remedy this problem is to break through the traditional egocentric constrain.

In this paper, we propose a novel decentralized multi-view algorithm, denoted as X-view. Inspired by previous methods, X-view fuses features of different coordinate spaces. In contrast to previous works, X-view breaks through the traditional ego-centric constraints that the origin of perspective view is consistent with the birds-eye view, limiting the power of multi-view fusion. X-view extends the number of perspective views by translating the origin to imitate different observers' perspectives. And \textit{X}, the number of PVs can be set flexibly. The intuition is that different origins in perspective view allow the network to exploit features in different distances with a dynamic context. Moreover, we propose BEV-Dominant Linear Interpolation Fusion(BDLI-Fusion) module to fuse features from multiple views. BDLI-Fusion fixes the issue existing in previous multi-view methods\cite{zhou2020end,wang2020pillar}. BDLI-Fusion avoids inaccuracies when conducting voxel-point-voxel projecting. The features from multiple views boost the final performance. And thanks to the efficient backbone architectures like 3D-conv\cite{graham2017submanifold}, PointNet++\cite{qi2017pointnet++}, the additional feature extraction brings little extra time and memory costs, and our X-view can achieve robust performance while guaranteeing the real-time demand. 


Our contributions can be summarized in three aspects: 

\begin{itemize}

\item We propose a novel multi-view 3D detection architectures, X-view. For the first time, X-view jumps out from the traditional egocentric view and extends the number of perspective views by translating the origin to imitate different observers' perspectives. 
\item Aiming at the fusion issue existing in previous multi-view methods, we propose BDLI-Fusion to simplify the fusion operation and avoid the accuracies introduced during voxel-point-voxel projection.
\item X-view, designed as a general, flexible architecture, can be applied on almost all mainstream 3D detectors. We conduct experiments on the four popular state-of-the-art 3D detectors: SECOND\cite{yan2018second}, PointRCNN\cite{shi2019pointrcnn}, Part-A$^2$\cite{shi2020points}, PV-RCNN\cite{shi2020pv}. The experiments on two challenging datasets, KITTI\cite{geiger2015kitti} and NuScenes \cite{caesar2020nuscenes}, demonstrate the effectiveness of X-view and show remarkable improvements.
\item X-view achieves real-time detection to meet the practical engineering demand via parallelization acceleration.
\end{itemize}
\begin{figure*}[t]
  \centering
  \includegraphics[width=0.98\textwidth]{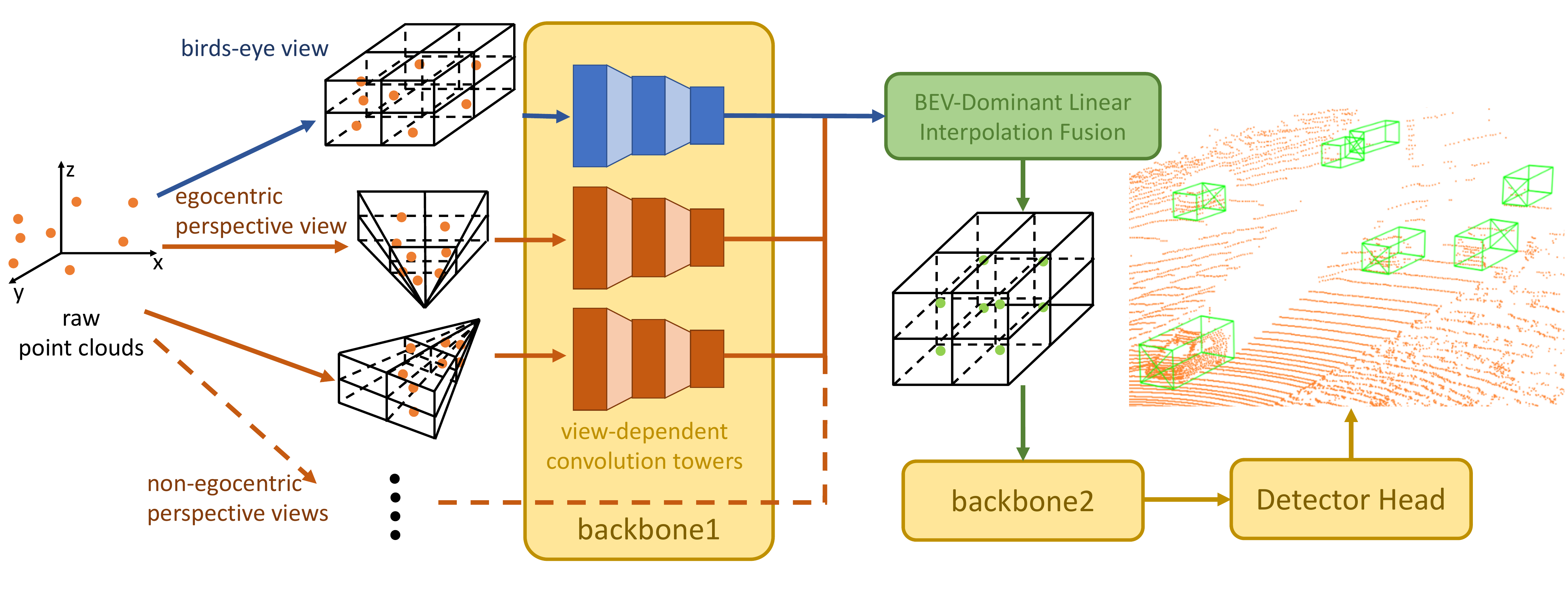}
  \caption{This figure illustrates the main architecture of X-view. Besides the birds-eye view and egocentric perspective view, X-view innovatively introduces the non-egocentric perspective views. X-view employs view-dependent backbones to extract features of each view(the \textit{backbone1} in the figure). Then, multi-view features are fused through our proposed \textit{BEV-Dominant Linear Interpolation Fusion} module. After fusion, the fused features are projected to BEV for further embedding in the \textit{backbone2} and finally to be feed into the \textit{detector head} to generate final predictions. \textbf{Note that the number of non-egocentric perspective views can be set flexibly according to the data scenes and demands.} The time cost brought by the multiple streams can be accelerated by parallel operation.}
  \label{fig:main_arch}
\end{figure*}

\section{Related work}
In this section, we review the recent development of 3D detection for autonomous driving. According to the number of views used by detectors, the 3D detection approaches can be grouped into two categories: single-view based and multi-view based methods. 


\subsection{Single-View 3D Detectors.} Some single-view 3D detectors\cite{li2019gs3d, manhardt2019roi, qin2019monogrnet, ku2019monocular, chang2019deep} employ monocular image to estimate 3D objects. However, due to the projective entanglement of depth and scale, precisely locating 3D targets is hard for the monocular 2D image. These approaches have a huge performance gap compared to LiDAR-based methods. LiDAR-based methods are mainly divided into two categories: grid-based and point-based. Grid-based detectors\cite{maturana2015voxnet, zhou2018voxelnet, yan2018second, zhu2019class, yang2018pixor, lang2019pointpillars, yin2020center} quantify the point clouds into a series of discrete grids and exploit efficient sparse 3D CNN\cite{zhou2018voxelnet, yan2018second, zhu2019class} or compress the input BEV feature maps\cite{yang2018pixor,lang2019pointpillars,yin2020center} to meet the real-time demand. Point-based methods directly process raw 3D points to aggregate point-level features via the popular PointNet/PointNet++ architecture\cite{shi2020points,qi2019deep,shi2020pv} or GCN\cite{wang2019dynamic}. However, although LiDAR-based 3D detectors recently achieve great successes, they only take advantage of points in the birds-eye view(\ie 3D Cartesian coordinate). Using points in the birds-eye view does not only suffer from the sparsity in the distance but also is unfriendly to the popular local-parameter-sharing feature extractors because the point densities vary with distance in the BEV. To remedy this issue, many researchers begin to handle point clouds in multiple views.

\subsection{Multi-View 3D Detectors.} For the multi-view paradigms, many recent works\cite{ku2018joint,chen2017multi,vora2019pointpainting,qi2020imvotenet} employ features from multiple sensors(\eg 3D LiDAR and 2D camera) to reason about 3D objects. Although these methods also use perspective view features, they only use 2D features. However, the 2D image does not contain sufficient 3D clues, making it hard to locate 3D objects precisely. And there are certain modality differences between features from the 2D image and 3D voxels or raw points, bringing difficulties for multi-sensor fusion. Hence, the performance of recent multi-sensor based methods still falls a little behind than LiDAR-only methods. Therefore, in the following, we mainly discuss the LiDAR-based multi-view methods, which are compatible with multi-sensor-based multi-view methods.

MV3D\cite{chen2017multi} is an early multi-view method. Except for the image stream, it exploits the point clouds in two views: birds-eye view and perspective view and directly fuses the feature maps of two views in the RoI-Pooling operation. There are two drawbacks. First, it projects the 3D points onto the 2D plane to represent the perspective view and loses the depth information, which harms the positioning precision. Second, it does not consider the inconsistency between the features maps from different perspectives and diametrically concatenates them together. MVF\cite{zhou2020end} maps the 3D \textit{xyz}-points into a 3D spherical coordinate to represent the perspective view. Meanwhile, it employs a dynamic point-to-voxel and voxel-to-point projection to conduct point-wise fusion to synergize the BEV and PV. However, because the sampling or pooling operation when mapping points to voxels brings some ambiguities, it reduces the fusion robustness. PillarOD\cite{wang2020pillar} exploits cylindrical projection to replace the spherical coordinate but still suffers the projection issue like MVF. Some methods\cite{meyer2019lasernet, chen2020every, liang2020rangercnn, rapoport2020s} leverage the range view or its variants, which can also be regarded as a kind of perspective view, to reason about 3D objects. CVCNet\cite{chen2020every} extracts the BEV and range-view features in a unified coordinate and fuses them via a cross-view transformer. RangeRCNN\cite{liang2020rangercnn} uses the range-view features as priors to initialize the BEV features to overcome the weakness of the range-image-based methods\cite{meyer2019lasernet}. \cite{rapoport2020s} uses features from cylindrical coordinate to guide the 3D convolution layer in cartesian coordinate. However, all of those multi-view methods only consider fusing features from the egocentric view, which makes the sparse feature in the distance mixed with the surrounding context and thus hampering the discriminability. To overcome those issues, we propose a novel decentralized multi-view fusion paradigm, X-view. X-view breaks through the concept of multiple views and extends the number of perspective views via imitating different observes' views to boost performance. Meanwhile, we propose BEV-Dominant Linear Interpolation Fusion(BDLI-Fusion) to remedy the projection fusion issue mentioned above. 

\section{X-View}

The main idea of X-view is breaking through the egocentric constraint for the perspective view via translating the origin of the perspective view coordinate and extending the number of perspective views to boost the performance.

The main architecture of X-view is illustrated in Figure \ref{fig:main_arch}. Besides the traditional Cartesian coordinate and the egocentric PV coordinate, the raw point clouds are also projected into series of non-egocentric perspective view coordinates. X-view applies a series of view-independent feature extractors. Then X-view employs the BEV-Dominant Linear Interpolation Fusion module (abbreviated as BDLI-Fusion) to fuse the features from multiple view streams. After the fusion, the fused multi-view features are feed into the following feature extractors to be further embedded and then are forwarded into the final detector head to generate the final predictions. In the following, we first introduce the concept of multiple views for the 3D detection. Then we introduce the detail of X-view and BDLI-Fusion, respectively.

\subsection{Multiple Views}\label{sec:multi_views}

In this section, we first compare the strengths and drawbacks of the traditional two views: birds-eye view and the egocentric perspective view, and then introduce our contribution: the non-egocentric perspective view.

\begin{figure}[t]
  \centering
  \includegraphics[width=0.48\textwidth]{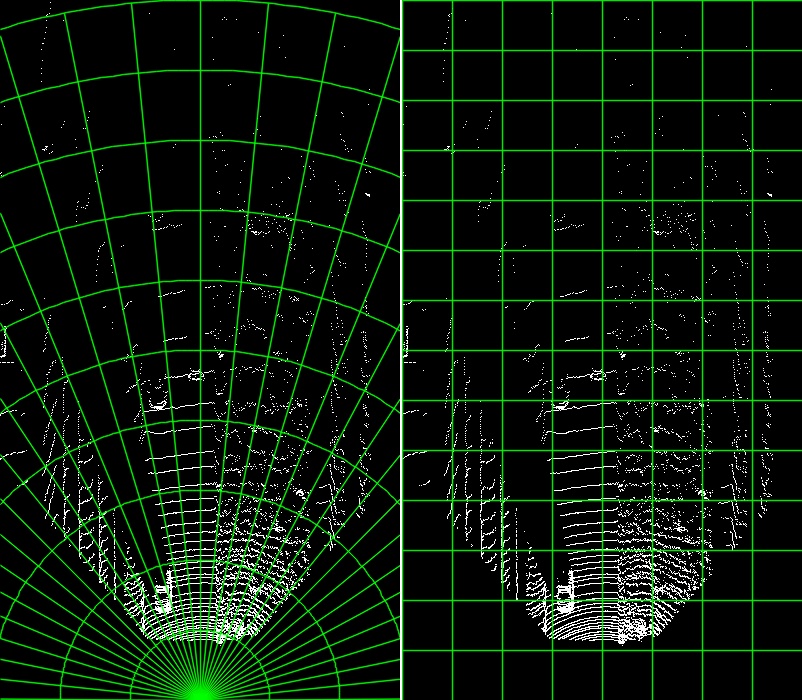}
  \caption{This figure compares the perspective view (spherical/cylindrical coordinate) and the birds-eye view (cartesian coordinate). The left is the perspective view, and the right is the birds-eye view. In the birds-eye view, the 3D space is divided into size-invariant voxels. However, the point density becomes small as the distance increases. The point densities in voxels are distance-variant, which causes the \textit{local feature mismatching} issue mentioned in Section \ref{sec:multi_views}. The perspective view eliminates this issue to some degree, and the point densities in the voxels are roughly consistent.}
  \label{fig:pv_bev}
\end{figure}

\textbf{Birds-eye View.} In this paper, we note the \textit{xyz}-cartesian-coordinate as birds-eye view(BEV). BEV performs very well in both grid-based methods\cite{ku2018joint,zhou2018voxelnet,yan2018second,shi2020points} and point-based methods\cite{shi2020point,wang2019dynamic,shi2019pointrcnn}, because its coordinate keep consistent with the measurements of evaluation metric. However, BEV also encounters some problems. Due to the working mechanism of LiDAR, although the laser beams emitted by LiDAR are dense and have consistent densities in angle, the point densities descent rapidly as the distance increases and the point densities of different local areas vary dramatically, as shown in the right part of Figure \ref{fig:pv_bev}. The unbalanced point density brings another problem: local feature mismatching. No matter the grid-based or voxel-based algorithms, they all leverage the local-parameter-sharing extractors to embed features, such as 2D/3D CNN or PointNet++. These local-parameter-sharing kernels will encounter problems when tackling the density-variant features in the BEV. Besides, the too-small point densities in the distances make detectors hard to recognize and locate the targets in the distance precisely. Another challenging case in the BEV is the objects of small sizes. For the BEV feature maps of deep level, the small targets might only occupy a few voxels or keypoints, making them hard to be precisely recognized and located.

\textbf{Perspective View.} Unlike the birds-eye view that captures the real 3D world, the perspective view(PV) imitates the perspective of the camera or the human eyes and describes a projection space. Although points are sparsely distributed in the 3D cartesian space, they will be dense when projected to the perspective-view-image like the process that LiDAR sweeps the scenes\cite{meyer2019lasernet}. Recent multi-view methods choose to use spherical\cite{zhou2020end} or cylindrical\cite{wang2020pillar} coordinate to represent perspective view. Compared to BEV, the voxel sizes of the PV coordinate vary with the distances from the origin, as illustrated in Figure \ref{fig:pv_bev}. Therefore, the point densities in the voxels are roughly consistent and can avoid the local feature mismatching issue mentioned above. Moreover, for the objects of small shapes, PV provides a "dense" description like in the 2D camera image. Although PV can remedy some drawbacks of the traditional BEV, it still suffers from the problems that distant voxels' sizes are too large. Specifically, the sparsity in the distance combined with surrounding clutter brought by the large voxel sizes hampers the discriminability of features.

\textit{Non-egocentric Perspective View.} In the PV coordinates, such as spherical or cylindrical coordinate, the voxel sizes become large as the distance to the origin increases, which causes PV is not robust for the distant targets. Therefore, our proposed X-view leverages the non-egocentric perspective view to remedying this issue. As the name suggests, the origin of non-egocentric PV is not located in the ego-car. To improve the robustness for those \textit{hard} objects in the distance, we can transform the origin of egocentric coordinate to the distant areas. As Figure \ref{fig:non-egocentric} illustrates, the non-egocentric PV can give a more meticulous division in the target area than the egocentric one. And the non-egocentric coordinate also imitates another observer in the scene.

\subsection{Non-egocentric Multi-View Detector}

X-view extends the concept of Multi-view based 3D detection. In previous multi-view methods, such as MVF\cite{zhou2020end} and PillarOD\cite{wang2020pillar}, the detector head relies on features extracted from two views: BEV and the egocentric PV. The origin of PV coordinate is consistent with the LiDAR coordinate. Although PV can remedy the local feature mismatching issue via its variant voxel sizes, it suffers from the large voxel shapes in the distance, making the features indiscriminative to precisely recognize and locate those distant targets. To remedy this issue, we propose X-view to break through the egocentric limitation and extends the number of perspective views to arbitrary number. The non-egocentric perspective views imitate multiple observers' perspectives, allowing us to leverage the characteristics of the perspective view to aggregate a balanced context across different distances. 

Considering the point clouds in Cartesian coordinates as $\mathcal{P}_{{\rm bev}}=\{(x_i, y_i, z_i)\}_{i=1}^N$ where $N$ is the number of the points, the traditional egocentric perspective view coordinate can be described as $\mathcal{P}^{\mathrm{ego}}_{{\rm pv}}=\{(r_i, \theta_i, \varphi_i)\}_{i=1}^N$, where:

\begin{equation}
	 r_i = \begin{cases}
            	\sqrt{x_i^2 + y_i^2 + z_i^2},  & \text{sphe coord.} \\[2ex]
            	\sqrt{x_i^2 + y_i^2},  & \text{cylin coord.,} \\
           \end{cases} \\
\end{equation}

\begin{equation}
	 \theta_i = \arctan\left({\frac{y_i}{x_i}}\right) \\
\end{equation}

\begin{equation}
	 \varphi_i = \begin{cases}
            		\arccos\left(\frac{z_i}{x_i^2 + y_i^2 + z_i^2}\right),  & \text{sphe coord.} \\[2ex]
            		z_i,  & \text{cylin coord.} \\
        		 \end{cases}
\end{equation}

where \textit{sphe coord.} and \textit{cylin coord.} represents the spherical and cylindrical coordinate respectively.

Non-egocentric perspective view coordinate translates the origin of egocentric one. Denote the non-egocentric perspective view centered at $(x_p, y_p, \varphi_p)$ as $\mathcal{P}_{{\rm pv}}^{(x_p, y_p, \varphi_p)} = \{(r_i, \theta_i, \varphi_i)\}_{(x_p, y_p, \varphi_p)}$, where:

\begin{equation}
	 r_i = \begin{cases}
            	\sqrt{(x_i - x_p)^2 + (y_i - y_p)^2 + (z_i - z_p)^2},  & \text{sphe coord.} \\[2ex]
            	\sqrt{(x_i - x_p)^2 + (y_i - y_p)^2},  & \text{cylin coord.} \\
           \end{cases} \\
\end{equation}

\begin{equation}
    \theta_i = \arctan\left({\frac{y_i - y_p}{x_i - x_p}}\right) 
\end{equation}

\begin{equation}
	 \varphi_i = \begin{cases}
            		\arccos\left(\frac{z_i - z_p}{(x_i - x_p)^2 + (y_i - y_p)^2 + (z_i - z_p)^2}\right),  & \text{sphe coord.} \\[2ex]
            		z_i - z_p,  & \text{cylin coord.} \\
        		 \end{cases}
\end{equation}


X-view considers both birds-eye view and a set of perspective view $\mathcal{P}_{{\rm pvs}} = \{\mathcal{P}_{{\rm pv}}^{(x_j, y_j, z_j)}\}_{j=1}^{K}$ centered at different location. The choice of the perspective centers depends on the LiDAR type and the road scenes, which thus differs for different datasets to fulfill different demands. In the Section \ref{exp:x-center}, we analysis the effects of different PV origin positions.



\subsection{BEV-Dominant Linear Interpolation Fusion}

\begin{figure}[t]
  \centering
  \includegraphics[width=0.48\textwidth]{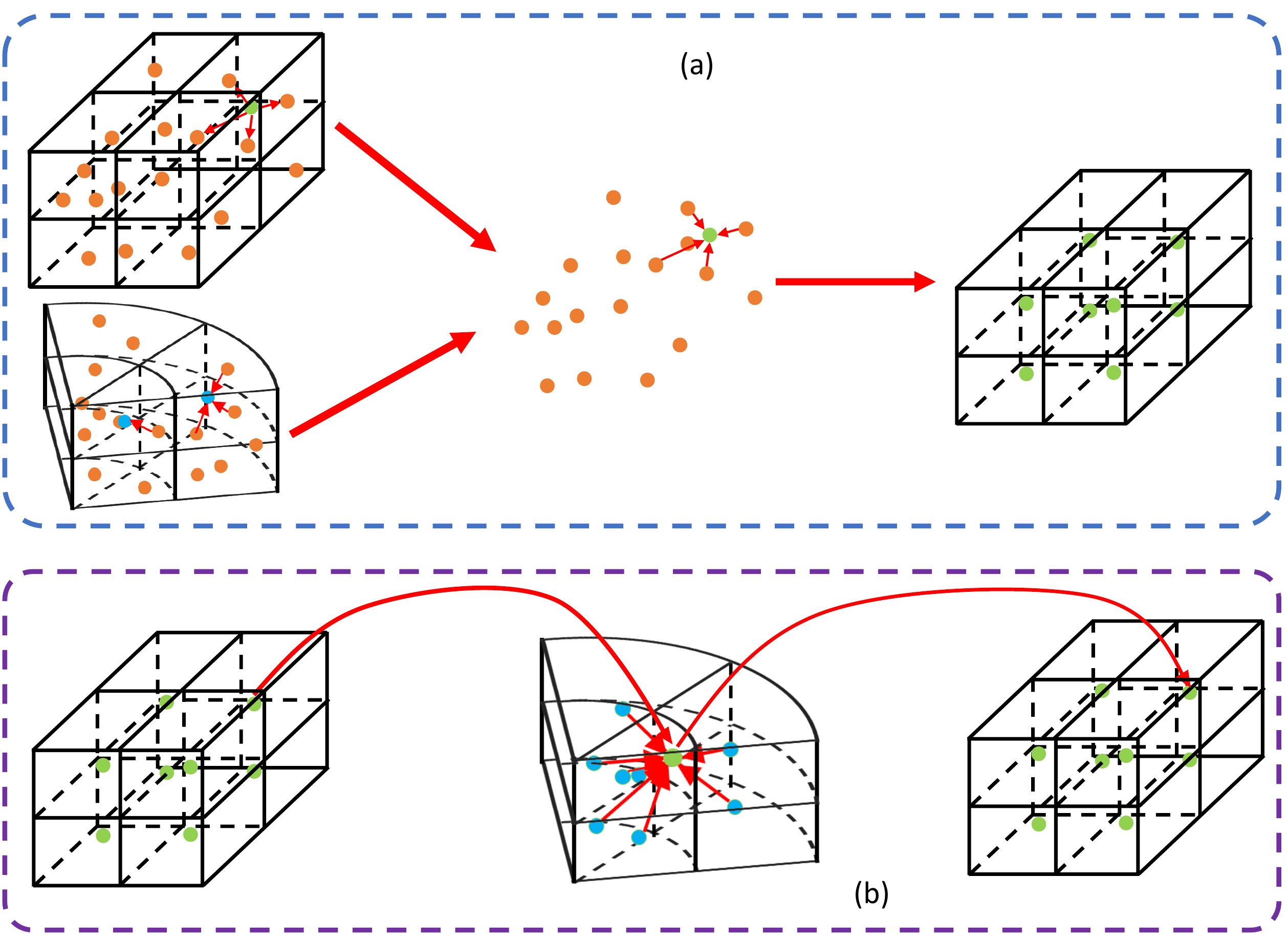}
  \caption{This figure illustrates the difference between the fusion procedure of X-view and the fusion of MVF\cite{zhou2020end} and PillarOD\cite{wang2020pillar}. The part (a) is depicts the fusion operation of MVF and PillarOD. They first retrieve the corresponding point features from the voxel feature maps of two views via the pre-constructed point-to-voxel mapping. After concatenating the features, they employ the voxel-to-point mapping to construct the voxel maps. However, this fusion procedure will involve some inaccuracy for the perspective view features. Although they using linear interpolation to retrieve point features from perspective view voxels, in the process of converting the points to birds-eye view voxels, the point position information with respect to the voxel centers will lose, which brings information loss. Therefore we remedy this issue as illustrated in part (b). We keep the birds-eye view voxels unchanged, and only project the birds-eye view voxel indices to the perspective view coordinate and retrieve the perspective view features via linear interpolation operation. Finally, we directly append the retrieved features onto the birds-eye view voxel features.}
  \label{fig:fusion}
\end{figure}


Previous multi-view methods \cite{zhou2020end, wang2020pillar} take points as a bridge to fuse features from different views, as shown in part (a) of the Figure \ref{fig:fusion}. However, we notice that this fusion operation will bring some inaccuracy. For the features in the birds-eye view, when retrieving point features from voxels, we use the linear interpolation. However, we loss the point relative positions with respect to the voxel center when recovering birds-eye view voxels. And the same issue exists for the perspective view. Since the voxels in birds-eye view(BEV) and in perspective view(PV) are voxelized in different coordinate spaces and their sizes are also inconsistent, the voxel(PV)-point-voxel(BEV) projection will also cause information losses, which are not beneficial for the following detection head. As Figure \ref{fig:fusion} explains, the problem happens at the point-to-voxel(BEV) mapping. When constructing voxel feature maps from point-wise features, for the points in the same voxel, the difference of position with respect to the voxel center loses, which causes that the points features blurs in the same voxel due to the sampling or pooling operation when building voxel feature maps. These inaccuracies harm the location precision especially the voxel sizes is large.

To remedy this issue, we propose BEV-Dominant Linear Interpolation Fusion, abbreviated as BDLI-Fusion. As illustrated in the part (b) of the Figure \ref{fig:fusion}, the detail process of BDLI-Fusion can be divided into three steps: (1) for each voxel in the birds-eye view, we project the \textit{xyz}-coordinate of the voxel centers to the perspective coordinate; (2) use linear interpolation to retrieve corresponding perspective view voxels; (3) fuse(concatenate or add operation) the retrieved features and birds-eye view voxel features together. Comparing the fusion operation of \cite{zhou2020end, wang2020pillar}, the BEV voxel features keep unchanged and consistent in the whole process of BDLI-Fusion. Because the following network layers and the detector head rely on the BEV voxels to reason about the final predictions, we directly use the BEV voxel centers as the fusion target to avoid the inaccuracy mentioned above. Compared to the two-pass fusion operation in MVF\cite{zhou2020end} and PillarOD\cite{wang2020pillar}, our BDLI-Fusion is more simple and efficient with a one-pass structure.

Note that our BDLI-Fusion is only an improved fusion for the architectures with grid-based detection head, such as \cite{zhou2018voxelnet,yan2018second,yang2018pixor,lang2019pointpillars,shi2020points,shi2020pv}. For those baselines with point-based detection head\cite{shi2019pointrcnn}, the fusion operation is more simple that we can directly fuse the retrieved point-wise features.

\subsection{Parallelization via Group Convolution}

\begin{figure}[t]
  \centering
  \includegraphics[width=0.48\textwidth]{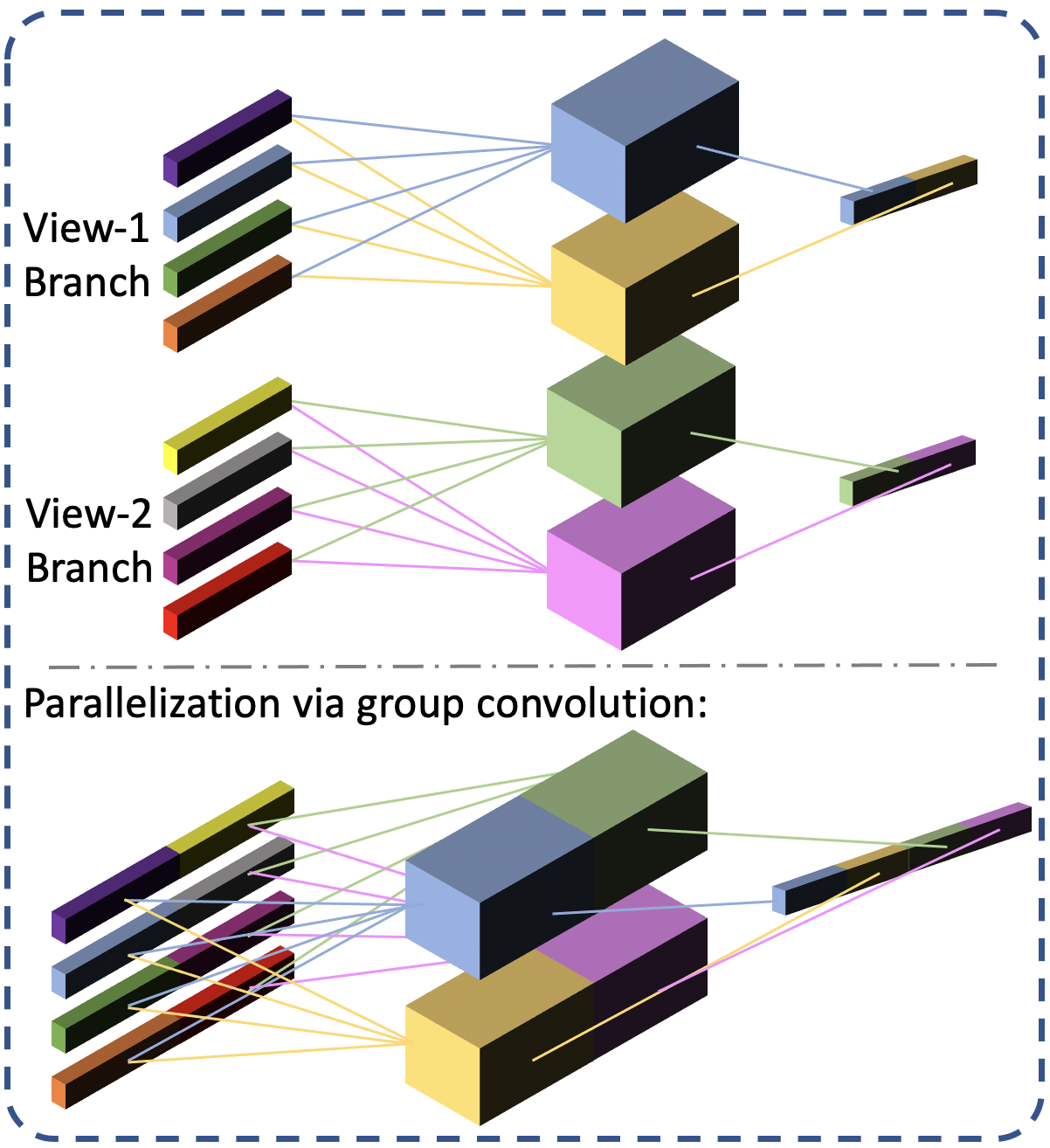}
  \caption{Group convolution is applied to parallelize the logically parallel network branches in the practical implementation. For normal implementation, we use \textit{for} loop to forward features of different views iteratively. Group convolution can take advantage of parallel computation and avoid the linear increment for the running speed. In the figure, the colors are corresponding in the upper and lower parts.}
  \label{fig:parallel}
\end{figure}

Real-time running time is another challenge that must be considered for 3D detection algorithms. Considering that the context differs in the different views, X-view applies view-dependent feature extractors for different views, as shown in Figure \ref{fig:main_arch}. Although the feature extractors of different views are logically parallel, the implementation via main-stream deep learning frameworks, such as \textit{PyTorch}, have to be designed as a serial computation graph architecture by using a \textit{for} loop. The serial computing procedure causes a linear increment in the running speed and makes it difficult to satisfy the real-time requirement. Considering that the number of input feature channels and the architectures of different view feature extractors are consistent, we can employ the group convolution operation to parallelize the computation to reduce the linear increment of running time. Figure \ref{fig:parallel} illustrates the parallelization via group convolution operation.

%

\section{Experiments}\label{sec:exp}

\begin{table*}[]
\caption{The comparison with four state-of-the-art methods: SECOND\cite{yan2018second}, PointRCNN\cite{shi2019pointrcnn}, Part-A$^2$\cite{shi2020points} and PV-RCNN\cite{shi2020pv}. The mmAP is the mean of the mAP of three categories: \textit{Car}, \textit{Pedestrian} and \textit{Cyclist}. \textbf{Note the results come from \textit{all-in-one} training procedure, \ie all models are trained for all three categories in one model.} The results are evaluated on the KITTI \textit{val} set. }
\centering
\label{exp:kitti_val_results}

\scalebox{1.0}{
\begin{tabular}{c|cccc|cccc}
\hline
Metrics        & \multicolumn{4}{c|}{3D mmAP}                                      & \multicolumn{4}{c}{BEV mmAP}   		                             \\ \hline
Methods        & SECOND\cite{yan2018second}         & PointRCNN\cite{shi2019pointrcnn}      & Part-A$^2$\cite{shi2020points}         & PV-RCNN\cite{shi2020pv}        & SECOND\cite{yan2018second}         & PointRCNN\cite{shi2019pointrcnn}      & Part-A$^2$\cite{shi2020points}         & PV-RCNN\cite{shi2020pv}          \\ \hline
baseline       & 67.51          & 72.78          & 73.00          & 74.22          & 72.18          & 76.25          & 77.53          & 77.49            \\
+egocentric PV & 68.32          & 72.62          & 73.72          & 74.80          & 73.41          & 76.89          & 77.03          & 78.42            \\
X-view(2 PVs) & \textbf{69.44} & \textbf{73.62} & \textbf{74.91} & \textbf{75.45} & \textbf{74.84} & \textbf{77.77} & \textbf{78.56} & \textbf{78.91}   \\ \hline
Improvements   & \textbf{+1.12} & \textbf{+1.00} & \textbf{+1.19} & \textbf{+0.65} & \textbf{+1.43} & \textbf{+0.88} & \textbf{+1.53} & \textbf{+0.49}   \\ \hline
\end{tabular}
}
\end{table*}


\begin{table*}[]
\caption{The detailed results on KITTI \textit{val} set. In this table, we choose SECOND\cite{yan2018second}, PointRCNN\cite{shi2019pointrcnn} and PV-RCNN\cite{shi2020pv} to show the detailed improvements on the \textit{Easy}, \textit{Moderate}, and \textit{Hard} targets of each category. In the table, the normal numbers represents the results of \textit{baseline+egocentric PV} and the numbers in the brackets represents the improvements of \textit{X-view(2PVs)} with respect to the number out of the brackets.}
\centering
\label{exp:kitti_detail_val_results}
\scalebox{0.94}{
\begin{tabular}{c|ccc|ccc|ccc}
\hline
Methods    & \multicolumn{3}{c|}{SECOND\cite{yan2018second}} & \multicolumn{3}{c|}{PointRCNN\cite{shi2019pointrcnn}} & \multicolumn{3}{c}{PV-RCNN\cite{shi2020pv}} \\ \hline
Category   & Easy   & Moderate   & Hard  & Easy    & Moderate    & Hard    & Easy   & Moderate   & Hard   \\ \hline
Car        & 88.54(+1.51) & 80.93(+0.30) & 77.97(+0.36) & 90.87(-0.17) & 79.96(+0.31) & 77.52(+0.36) & 91.65(+0.36) & 84.39(+0.24) & 82.03(+0.53) \\
Pedestrian & 56.93(+2.10) & 51.66(+1.78) & 47.09(+1.04) & 65.43(+1.21) & 58.17(-0.07) & 51.01(+0.54) & 67.47(+0.19) & 60.56(-0.15) & 55.68(-0.39) \\
Cyclist    & 83.13(-0.51) & 66.08(+2.12) & 62.49(+1.45) & 92.21(+1.92) & 71.45(+2.64) & 66.93(+2.27) & 66.93(+2.27) & 72.62(+1.51) & 68.05(+1.75) \\ \hline
\end{tabular}
}
\end{table*}

\begin{table*}[]
\caption{The results on NuScenes \textit{val} set. The baseline models is SECOND\cite{yan2018second}. In the Table, \textit{Ego PV} represents the egocentric PV and \textit{X-view} use 1 egocentric PV and 2 non-egocentric PVs whose origins are (40$m$, 0$m$, 0$m$) and (-40$m$, 0$m$, 0$m$) respectively.}
\centering
\label{exp:nuscenes}
\scalebox{1.0}{
\begin{tabular}{c|cccccccccc|cc}
\hline
 Methods &
  car &
  truck &
  bus &
  trailer &
  \begin{tabular}[c]{@{}c@{}}pedes-\\ trian\end{tabular} &
  \begin{tabular}[c]{@{}c@{}}motor-\\ cycle\end{tabular} &
  bicycle &
  \begin{tabular}[c]{@{}c@{}}traffic-\\ cone\end{tabular} &
  \begin{tabular}[c]{@{}c@{}}constru-\\ ction-\\ vehicle\end{tabular} &
  barrier &
  mAP &
  NDS \\ \hline
baseline          & 81.48 & 51.37 & 66.65 & 37.21 & 76.98 & 43.01 & 16.88 & 57.12 & 14.68 & 59.14 & 50.45 & 61.99 \\
baseline + Ego PV & 82.01 & 51.82 & 66.68 & 38.12 & 77.54 & 43.16 & 17.07 & 58.23 & 14.98 & 59.97 & 50.96 & 62.87 \\
X-view (3 PVs)    & \textbf{83.45} & \textbf{53.24} & \textbf{67.33} & \textbf{38.24} & \textbf{78.56} & \textbf{43.25} & \textbf{17.74} & \textbf{58.89} & \textbf{15.11} & \textbf{61.03} & \textbf{51.69} & \textbf{63.76} \\ \hline
\end{tabular}
}
\end{table*}

\begin{table}[]
\caption{The results on KITTI \textit{test} set. We compare X-view with Part-A$^2$\cite{shi2020points}. The baseline results are official public results on the \textit{KITTI 3D Object Benchmark}.}
\centering
\label{exp:kitti_test_results}
\scalebox{1.0}{
\begin{tabular}{c|c|ccc|l}
\hline
\multicolumn{2}{c|}{Method}                          & Easy  & Moderate & Hard  & mAP   \\ \hline
\multirow{2}{*}{Part-A\textasciicircum{}2} & baseline & \textbf{87.81} & 78.49    & 73.51 & 79.94 \\
                                          & X-view   & 87.72 & \textbf{80.41}    & \textbf{76.22} & \textbf{81.45} \\ \hline
\multirow{2}{*}{PV-RCNN}                  & baseline & \textbf{90.25} & \textbf{81.43} 	& 76.82 & \textbf{82.83} \\
                                          & X-view   & 89.21 & 81.35    & \textbf{76.87} &              82.47 \\ \hline
\end{tabular}
}
\end{table}

In this section, we first introduce the settings of the experiment in the Section \ref{sec:exp_settings}; then we compare our X-view and recent state-of-the-art methods in the Section \ref{sec:results}; finally, we conduct some ablation studies to analyze the performance of X-view in the Section \ref{sec:ablation}.

\subsection{Settings}\label{sec:exp_settings}

\subsubsection{Datasets}

We compare our proposed X-view on two challenging public datasets: KITTI\cite{geiger2015kitti} and NuScenes\cite{caesar2020nuscenes}.

\textbf{KITTI Dataset.} KITTI\cite{geiger2015kitti} is a popular 3D object detection dataset for autonomous driving. It contains 7481 samples with labels for training and validating and 7518 labels without labels for testing. For 3D object detection, each sample provides a frame of LiDAR point clouds, an RGB image, and the calibration information. The training samples are labeled with a list of the ground-truth 3D bounding boxes and the corresponding 2D bounding boxes on the 2D image. The ground-truth only contains the objects in the 2D image. Hence, when training and evaluating, we only input the 3D points in the camera frustum as the popular practices. When conducting experiments, we follow the popular practice like\cite{chen2017multi} to divide the samples with labels into two sets: 3712 samples as \textit{train} set and 3769 samples as \textit{val} set. 

\textbf{NuScenes Dataset.} Compared to KITTI dataset, NuScenes\cite{caesar2020nuscenes} dataset is a more challenging dataset of 3D object detection for autonomous driving. NuScenes comprises of 1000 scenes with 10 classes, which are then divided into 700 scenes for training and 150 scenes for validation, and 150 scenes for testing. Specifically, each scene is a series of consecutive frames of 20$s$ duration, and annotated keyframes are sample at the frequency of 2Hz. And the consecutive frames with calibration makes it possible to use multiple LiDAR sweeps to enhance the single-frame point clouds. Totally, it has 28000, 6000, 6000 annotated keyframes for training, validation and testing, respectively. 

\subsubsection{Experiments Details}

X-view is implemented based on the official repository of PV-RCNN\cite{shi2020pv} on GitHub. We employ consistent configurations for each baseline model when training and testing. On the KITTI dataset, we train the model with the batch size of 4, the initial learning rate of 0.003, weight decay of 0.01, the momentum of 0.9 on a single TITAN RTX GPU for 80 epochs. On the NuScenes dataset, we train the model with the batch size of 8, the initial learning rate of 0.003, weight decay of 0.01, the momentum of 0.9 on a single TITAN RTX GPU. We train the model for 20 epochs on the \textit{train} split. On both two datasets, all models are end-to-end trained from scratch and apply the ADAM optimizer to update model parameters. \textbf{All our experiments, we employ \textit{all-in-one} training including the results evaluating on the \textit{test} set, \ie we train all categories in one model, which is different from the common practice of previous works.} Therefore, to compare fairly, we reimplement the baselines in all experiments and only compare the improvements to our reimplemented baselines.

\textbf{Data Augmentation.} We follow the commonly adopted data augmentations including randomly scaling with a random factor sampled between [0.95, 1.05], randomly flipping along the \textit{x} axis, randomly rotating around the \textit{z} axis with a rotation angle randomly sampled between [$-\pi/4, \pi/4$]. We also employ the popular \textit{gt-aug}\cite{yan2018second} to randomly place new the ground-truths from other samples to increase the number of positive samples and the complexity of the scenes. 

\textbf{Evaluation Metrics.} For KITTI dataset, we evaluate all models on the metrics of BEV and 3D AP(Average Precision). The AP is calculated with 40 recall positions. For NuScenes dataset, we test on the metrics of mAP and NDS(nuScenes detection score). The mAP is based on the distance threshold: 0.5$m$, 1.0$m$, 2.0$m$ and 4.0$m$, and the NDS is a weighted sum of mAP and precision on box location, scale, orientation, velocity and attributes.

\subsection{Results}\label{sec:results}

\subsubsection{Results on KITTI datasets}

Table \ref{exp:kitti_val_results} shows the comparisons among X-view and four state-of-the-art baselines: SECOND\cite{yan2018second}, PointRCNN\cite{shi2019pointrcnn}, Part-A$^2$\cite{shi2020points} and PV-RCNN\cite{shi2020pv}. The mmAP in the table is the mean of the mAPs (\ie the mean of the APs for \textit{Easy}, \textit{Moderate} and \textit{Hard} diffculties) for three categories: \textit{Car}, \textit{Pedestrian} and \textit{Cyclist}. In the table,  \textit{+egocentric PV} means adding a egocentric PV(perspective view) based on the baseline. \textit{X-view(2 PVs)} means leveraging a non-egocentric PV stream besides the egocentric one whose origin is at (60$m$, 0$m$, 0$m$). The results shows that X-view can obtain remarkable improvements compared to both the baseline and the baseline with egocentric PV on both 3D and BEV AP metrics.  In the Table \ref{exp:kitti_detail_val_results}, we display the detailed improvements on the \textit{Easy}, \textit{Moderate}, and \textit{Hard} objects of three categories. The results demonstrates that X-view can achieve significant improvements on almost all categories and diffculty levels.


Table \ref{exp:kitti_test_results} shows the performance comparison on the KITTI \textit{test} set. We compare our X-view with the state-of-the-art methods, Part-A$^2$\cite{shi2020points}. The \textit{baseline} results are the public results on the official \textit{KITTI 3D Object Benchmark}. Through the results, we can see that our X-view can achieve improvements on the metric of mAP, and the improvement on \textit{Moderate} and \textit{Hard} objects is especially remarkable. The improvements on the distant objects demonstrate the non-egocentric PV of X-view can fix the large-voxel-size issue of egocentric view in the distant area.

\begin{table}[]
\caption{The ablation study of the performance differences between the fusion methods of MVF and our proposed BDLI-Fusion. The results in this table are the 3D mAP for three categories.}
\centering
\label{exp:fusion}
\scalebox{1.0}{
\begin{tabular}{c|ccc|c}
\hline
fusion method &  Car  & Pedestrian & Cyclist & mmAP  \\ \hline
Old Fusion    & 83.13 &   52.76    &  70.62 	 & 68.83 \\
BDLI-Fusion   & \textbf{83.20} &   \textbf{53.53}    &  \textbf{71.59}  & \textbf{69.44} \\ \hline
\end{tabular}
}
\end{table}

\begin{table}[]
\caption{Study about the number of the non-egocentric PVs. The results are evaluated on the metric of 3D AP for \textit{Car} category. "non-ego PV" represents the non-egocentric perspective views.}
\centering
\label{exp:num_pv}
\scalebox{0.88}{
\begin{tabular}{c|ccc|c|c|c}
\hline
\begin{tabular}[c]{@{}c@{}}number of \\ non-ego PVs\end{tabular} &
  Easy &
  Moderate &
  Hard &
  mAP &
  \begin{tabular}[c]{@{}c@{}}running\\ time(ms)\end{tabular} &
  \begin{tabular}[c]{@{}c@{}}GPU\\ memory(GB)\end{tabular} \\ \hline
1 & 93.12 & 75.98 & 72.38 & 80.49 & 58 & 2.07 \\
2 & 91.91 & 77.52 & 73.91 & 81.11 & 71 & 2.97 \\
3 & 93.35 & 77.64 & 72.85 & 81.28 & 80 & 3.25 \\ \hline
\end{tabular}
}
\end{table}

\begin{table}[]
\caption{Ablation study about the performance difference between \textit{spherical} and \textit{cylindrical} coordinates. "Ego PV" represents the egocentric perspective view. "X-view" applies one more non-egocentric perspective view than the "baseline + Ego PV".}
\centering
\label{exp:sph_cyc}
\scalebox{0.96}{
\begin{tabular}{c|c|ccc|c}
\hline
Model & Coordinate  & Easy  & Moderate & Hard & mAP \\ \hline
\multirow{2}{*}{\begin{tabular}[c]{@{}c@{}}baseline + Ego PV \end{tabular}} & spherical & 90.76 & 76.10 & 72.36 & 79.74 \\
      & cylindrical & 92.34 & 76.57 & 68.47 & 79.46 \\ \hline
\multirow{2}{*}{\begin{tabular}[c]{@{}c@{}}X-view \end{tabular}}          & spherical & 93.12 & 75.98 & 72.38 & 80.49 \\
      & cylindrical & 91.15 & 76.04 & 72.82 & 80.01 \\ \hline
\end{tabular}
}
\end{table}

\begin{table}[]
\caption{The ablation study about the origin position of the perspective view. The results are 3D AP for \textit{Car} category. }
\centering
\label{exp:x-center}
\scalebox{1.0}{
\begin{tabular}{c|ccc|c}
\hline
origin       & Easy  & Moderate & Hard  & mean  \\ \hline
(20, 0, 0)   & 91.74 & 75.42    & 72.03 & 79.73 \\
(40, 0, 0)   & \textbf{93.12} & 75.98    & 72.38 & 80.49 \\
(60, 0, 0)   & 92.45 & \textbf{77.95}    & \textbf{74.06} & \textbf{81.49} \\
(60, -20, 0) & 91.68 & 76.68    & 72.81 & 80.39 \\
(60, 20, 0)  & 90.85 & 76.74    & 73.01 & 80.20 \\ \hline
\end{tabular}
}
\end{table}

\subsubsection{Results on NuScenes dataset}

Table \ref{exp:nuscenes} shows the comparison between X-view and baseline on the \textit{val} set of NuScenes\cite{caesar2020nuscenes} dataset. The baseline model is SECOND\cite{yan2018second}. The point clouds in the NuScenes dataset are 360-degree scanned by LiDAR, and annotations contain the objects behind the ego car. Therefore, we place one more non-egocentric PV compared to the setting on the KITTI dataset. The origins of non-egocentric PVs of the 3-rd row are set as (40$m$, 0$m$, 0$m$) and (-40$m$, 0$m$, 0$m$) respectively. The results on NuScenes suggests that our X-view can achieve improvements compared to the baseline with only egocentric PV like on the KITTI dataset. 

\subsection{Ablation Studies}\label{sec:ablation}

We do some ablation analysis on the KITTI datasets to study each design of X-view. All ablation studies are trained on the KITTI \textit{train} set and evaluated on the \textit{val} set, and the baseline model is SECOND\cite{yan2018second}.


\textbf{Number of Non-egocentric Perspective Views.} One of X-view's advantages is that the number of non-egocentric PV(perspective views) is extendable. Therefore, we investigate the effect of the number of non-egocentric PV in Table \ref{exp:num_pv}. In the table, the first row "1 non-ego PV" uses the origin point in the position of (40$m$, 0$m$, 0$m$); the 2-nd row is (40$m$, -20$m$, 0$m$) and (40$m$, 20$m$, 0$m$); the 3-rd row is (60$m$, -20$m$, 0$m$), (40$m$, -20$m$, 0$m$) and (40$m$, 20$m$, 0$m$). Through the results, we can see that as the number of PVs increases, the performance will also be better, but the running speed and the GPU memory also go up. We can find that the marginal increment of the performance reduces as the number of PVs increases. Therefore, we have to find a trade-off point between the number of PVs and speed. And the results of Table \ref{exp:num_pv} only provide advice for the KITTI dataset (3D points in the KITTI dataset are valid in the front camera frustum). The origin positions and the number of non-egocentric PVs should be adjusted according to the dataset and road scenes.

\textbf{Ablation study of the Fusion Methods.} In Table \ref{exp:fusion}, we compare the effects of different fusion methods: the fusion of MVF\cite{zhou2020end} and our BDLI-Fusion. We can see that the BDLI-Fusion can bring improvements through the results, especially for the \textit{Cyclist} and \textit{Pedestrian} categories. It is rational because the object sizes of these two categories are generally small, and the estimation for these objects are easy to be influenced by the inaccurate features.

\textbf{Spherical or Cylindrical Coordinate?} \cite{wang2020pillar} shows that the cylindrical coordinate can avoid the distort of the objects' physical scale compared to the spherical coordinate. However, this is only for the case of pillar-based baseline\cite{lang2019pointpillars}, because the pillar-based methods do not voxelize the \textit{z} axis direction. Many grid-based backbones voxelize all \textit{x}, \textit{y}, \textit{z} directions to 3D voxels. For these methods, applying a cylindrical coordinate will face the density-inconsistent issue in the \textit{z} axis. So we reinvestigate the performance difference between the \textit{spherical} and the \textit{cylindrical} coordinates In Table \ref{exp:sph_cyc}. It suggests that no matter for baseline+PV model or our X-view model, although spherical coordinate is little better than cylindrical, different coordinates do not introduce a significant difference. 

\textbf{Where to Place X-view?} We investigate the effects of the different origin locations of non-egocentric PV on the detection performance. As shown in the first to third rows, the performance on \textit{Moderate} and \textit{Hard} difficulty levels increases as the origin slides away along the x-axis from near to far, while the performance of \textit{Easy} objects will peak at a point and then drop, which is reasonable as the best-mixed voxel size in the \textit{Easy} difficulty levels will peak faster than \textit{Moderate} and \textit{Hard} difficulty levels. The non-egocentric PV placed in the far area will provide a more detailed grid partition and more discriminative features for the distant targets. What's more, the 4-th to 6-th rows show that placing the origin at the symmetry axis helps us achieve the best detection performance. It is reasonable because the obstacles and road scenes are often symmetrically distributed with respect to the \textit{x} axis, \ie the driving direction, and the input are only the point clouds in the front camera frustum in the KITTI dataset.

\section{Conclusion}

In this paper, we generalize the research on multi-view 3D Object Detection. We propose a novel non-egocentric multi-view 3D detector, named X-view. X-view breaks through the traditional egocentric constraint that the origin of perspective view is consistent with ego coordinate. By adding the non-egocentric perspective views, X-view remedies the issue that the distant voxel sizes became too large in the egocentric view. Besides, X-view leverages our improved fusion module: BDLI Fusion, which fixes the point-wise fusion issue existing in previous methods. X-view architecture can achieve remarkable improvements on four state-of-the-art 3D detection methods: SECOND\cite{yan2018second}, PointRCNN\cite{shi2019pointrcnn}, Part-A$^2$\cite{shi2020points} and PV-RCNN\cite{shi2020pv}. And based on the parallelization operation, different view streams can be accelerated to avoid the linear increment of the running time, making X-view can meet the real-time demand.

\ifCLASSOPTIONcaptionsoff
  \newpage
\fi



\bibliographystyle{IEEEtran}
\bibliography{Bibliography}

\begin{thebibliography}{10}
\providecommand{\url}[1]{#1}
\csname url@samestyle\endcsname
\providecommand{\newblock}{\relax}
\providecommand{\bibinfo}[2]{#2}
\providecommand{\BIBentrySTDinterwordspacing}{\spaceskip=0pt\relax}
\providecommand{\BIBentryALTinterwordstretchfactor}{4}
\providecommand{\BIBentryALTinterwordspacing}{\spaceskip=\fontdimen2\font plus
\BIBentryALTinterwordstretchfactor\fontdimen3\font minus
  \fontdimen4\font\relax}
\providecommand{\BIBforeignlanguage}[2]{{%
\expandafter\ifx\csname l@#1\endcsname\relax
\typeout{** WARNING: IEEEtran.bst: No hyphenation pattern has been}%
\typeout{** loaded for the language `#1'. Using the pattern for}%
\typeout{** the default language instead.}%
\else
\language=\csname l@#1\endcsname
\fi
#2}}
\providecommand{\BIBdecl}{\relax}
\BIBdecl

\bibitem{geiger2015kitti}
A.~Geiger, P.~Lenz, C.~Stiller, and R.~Urtasun, ``The kitti vision benchmark
  suite,'' \emph{URL http://www. cvlibs. net/datasets/kitti}, 2015.

\bibitem{caesar2020nuscenes}
H.~Caesar, V.~Bankiti, A.~H. Lang, S.~Vora, V.~E. Liong, Q.~Xu, A.~Krishnan,
  Y.~Pan, G.~Baldan, and O.~Beijbom, ``nuscenes: A multimodal dataset for
  autonomous driving,'' in \emph{Proceedings of the IEEE/CVF Conference on
  Computer Vision and Pattern Recognition}, 2020, pp. 11\,621--11\,631.

\bibitem{yan2018second}
Y.~Yan, Y.~Mao, and B.~Li, ``Second: Sparsely embedded convolutional
  detection,'' \emph{Sensors}, vol.~18, no.~10, p. 3337, 2018.

\bibitem{shi2019pointrcnn}
S.~Shi, X.~Wang, and H.~Li, ``Pointrcnn: 3d object proposal generation and
  detection from point cloud,'' in \emph{Proceedings of the IEEE Conference on
  Computer Vision and Pattern Recognition}, 2019, pp. 770--779.

\bibitem{shi2020points}
S.~Shi, Z.~Wang, J.~Shi, X.~Wang, and H.~Li, ``From points to parts: 3d object
  detection from point cloud with part-aware and part-aggregation network,''
  \emph{IEEE Transactions on Pattern Analysis and Machine Intelligence}, 2020.

\bibitem{shi2020pv}
S.~Shi, C.~Guo, L.~Jiang, Z.~Wang, J.~Shi, X.~Wang, and H.~Li, ``Pv-rcnn:
  Point-voxel feature set abstraction for 3d object detection,'' in
  \emph{Proceedings of the IEEE/CVF Conference on Computer Vision and Pattern
  Recognition}, 2020, pp. 10\,529--10\,538.

\bibitem{zhou2018voxelnet}
Y.~Zhou and O.~Tuzel, ``Voxelnet: End-to-end learning for point cloud based 3d
  object detection,'' in \emph{Proceedings of the IEEE Conference on Computer
  Vision and Pattern Recognition}, 2018, pp. 4490--4499.

\bibitem{lang2019pointpillars}
A.~H. Lang, S.~Vora, H.~Caesar, L.~Zhou, J.~Yang, and O.~Beijbom,
  ``Pointpillars: Fast encoders for object detection from point clouds,'' in
  \emph{Proceedings of the IEEE Conference on Computer Vision and Pattern
  Recognition}, 2019, pp. 12\,697--12\,705.

\bibitem{shi2020point}
W.~Shi and R.~Rajkumar, ``Point-gnn: Graph neural network for 3d object
  detection in a point cloud,'' in \emph{Proceedings of the IEEE/CVF Conference
  on Computer Vision and Pattern Recognition}, 2020, pp. 1711--1719.

\bibitem{wang2019dynamic}
Y.~Wang, Y.~Sun, Z.~Liu, S.~E. Sarma, M.~M. Bronstein, and J.~M. Solomon,
  ``Dynamic graph cnn for learning on point clouds,'' \emph{Acm Transactions On
  Graphics (tog)}, vol.~38, no.~5, pp. 1--12, 2019.

\bibitem{meyer2019lasernet}
G.~P. Meyer, A.~Laddha, E.~Kee, C.~Vallespi-Gonzalez, and C.~K. Wellington,
  ``Lasernet: An efficient probabilistic 3d object detector for autonomous
  driving,'' in \emph{Proceedings of the IEEE Conference on Computer Vision and
  Pattern Recognition}, 2019, pp. 12\,677--12\,686.

\bibitem{qi2017pointnet++}
C.~R. Qi, L.~Yi, H.~Su, and L.~J. Guibas, ``Pointnet++: Deep hierarchical
  feature learning on point sets in a metric space,'' in \emph{Advances in
  neural information processing systems}, 2017, pp. 5099--5108.

\bibitem{chen2017multi}
X.~Chen, H.~Ma, J.~Wan, B.~Li, and T.~Xia, ``Multi-view 3d object detection
  network for autonomous driving,'' in \emph{Proceedings of the IEEE Conference
  on Computer Vision and Pattern Recognition}, 2017, pp. 1907--1915.

\bibitem{zhou2020end}
Y.~Zhou, P.~Sun, Y.~Zhang, D.~Anguelov, J.~Gao, T.~Ouyang, J.~Guo, J.~Ngiam,
  and V.~Vasudevan, ``End-to-end multi-view fusion for 3d object detection in
  lidar point clouds,'' in \emph{Conference on Robot Learning}, 2020, pp.
  923--932.

\bibitem{wang2020pillar}
Y.~Wang, A.~Fathi, A.~Kundu, D.~Ross, C.~Pantofaru, T.~Funkhouser, and
  J.~Solomon, ``Pillar-based object detection for autonomous driving,''
  \emph{arXiv preprint arXiv:2007.10323}, 2020.

\bibitem{graham2017submanifold}
B.~Graham and L.~van~der Maaten, ``Submanifold sparse convolutional networks,''
  \emph{arXiv preprint arXiv:1706.01307}, 2017.

\bibitem{li2019gs3d}
B.~Li, W.~Ouyang, L.~Sheng, X.~Zeng, and X.~Wang, ``Gs3d: An efficient 3d
  object detection framework for autonomous driving,'' in \emph{Proceedings of
  the IEEE Conference on Computer Vision and Pattern Recognition}, 2019, pp.
  1019--1028.

\bibitem{manhardt2019roi}
F.~Manhardt, W.~Kehl, and A.~Gaidon, ``Roi-10d: Monocular lifting of 2d
  detection to 6d pose and metric shape,'' in \emph{Proceedings of the IEEE
  Conference on Computer Vision and Pattern Recognition}, 2019, pp. 2069--2078.

\bibitem{qin2019monogrnet}
Z.~Qin, J.~Wang, and Y.~Lu, ``Monogrnet: A geometric reasoning network for
  monocular 3d object localization,'' in \emph{Proceedings of the AAAI
  Conference on Artificial Intelligence}, vol.~33, 2019, pp. 8851--8858.

\bibitem{ku2019monocular}
J.~Ku, A.~D. Pon, and S.~L. Waslander, ``Monocular 3d object detection
  leveraging accurate proposals and shape reconstruction,'' in
  \emph{Proceedings of the IEEE Conference on Computer Vision and Pattern
  Recognition}, 2019, pp. 11\,867--11\,876.

\bibitem{chang2019deep}
J.~Chang and G.~Wetzstein, ``Deep optics for monocular depth estimation and 3d
  object detection,'' in \emph{Proceedings of the IEEE International Conference
  on Computer Vision}, 2019, pp. 10\,193--10\,202.

\bibitem{maturana2015voxnet}
D.~Maturana and S.~Scherer, ``Voxnet: A 3d convolutional neural network for
  real-time object recognition,'' in \emph{2015 IEEE/RSJ International
  Conference on Intelligent Robots and Systems (IROS)}.\hskip 1em plus 0.5em
  minus 0.4em\relax IEEE, 2015, pp. 922--928.

\bibitem{zhu2019class}
B.~Zhu, Z.~Jiang, X.~Zhou, Z.~Li, and G.~Yu, ``Class-balanced grouping and
  sampling for point cloud 3d object detection,'' \emph{arXiv preprint
  arXiv:1908.09492}, 2019.

\bibitem{yang2018pixor}
B.~Yang, W.~Luo, and R.~Urtasun, ``Pixor: Real-time 3d object detection from
  point clouds,'' in \emph{Proceedings of the IEEE conference on Computer
  Vision and Pattern Recognition}, 2018, pp. 7652--7660.

\bibitem{yin2020center}
T.~Yin, X.~Zhou, and P.~Kr{\"a}henb{\"u}hl, ``Center-based 3d object detection
  and tracking,'' \emph{arXiv preprint arXiv:2006.11275}, 2020.

\bibitem{qi2019deep}
C.~R. Qi, O.~Litany, K.~He, and L.~J. Guibas, ``Deep hough voting for 3d object
  detection in point clouds,'' in \emph{Proceedings of the IEEE/CVF
  International Conference on Computer Vision}, 2019, pp. 9277--9286.

\bibitem{ku2018joint}
J.~Ku, M.~Mozifian, J.~Lee, A.~Harakeh, and S.~L. Waslander, ``Joint 3d
  proposal generation and object detection from view aggregation,'' in
  \emph{2018 IEEE/RSJ International Conference on Intelligent Robots and
  Systems (IROS)}.\hskip 1em plus 0.5em minus 0.4em\relax IEEE, 2018, pp. 1--8.

\bibitem{vora2019pointpainting}
S.~Vora, A.~H. Lang, B.~Helou, and O.~Beijbom, ``Pointpainting: Sequential
  fusion for 3d object detection,'' \emph{arXiv preprint arXiv:1911.10150},
  2019.

\bibitem{qi2020imvotenet}
C.~R. Qi, X.~Chen, O.~Litany, and L.~J. Guibas, ``Imvotenet: Boosting 3d object
  detection in point clouds with image votes,'' \emph{arXiv preprint
  arXiv:2001.10692}, 2020.

\bibitem{chen2020every}
Q.~Chen, L.~Sun, E.~Cheung, and A.~L. Yuille, ``Every view counts: Cross-view
  consistency in 3d object detection with hybrid-cylindrical-spherical
  voxelization,'' \emph{Advances in Neural Information Processing Systems},
  vol.~33, 2020.

\bibitem{liang2020rangercnn}
Z.~Liang, M.~Zhang, Z.~Zhang, X.~Zhao, and S.~Pu, ``Rangercnn: Towards fast and
  accurate 3d object detection with range image representation,'' \emph{arXiv
  preprint arXiv:2009.00206}, 2020.

\bibitem{rapoport2020s}
M.~Rapoport-Lavie and D.~Raviv, ``It's all around you: Range-guided cylindrical
  network for 3d object detection,'' \emph{arXiv preprint arXiv:2012.03121},
  2020.

\end{thebibliography}
%



%

\begin{IEEEbiography}[{\includegraphics[width=1in,height=1.25in,clip,keepaspectratio]{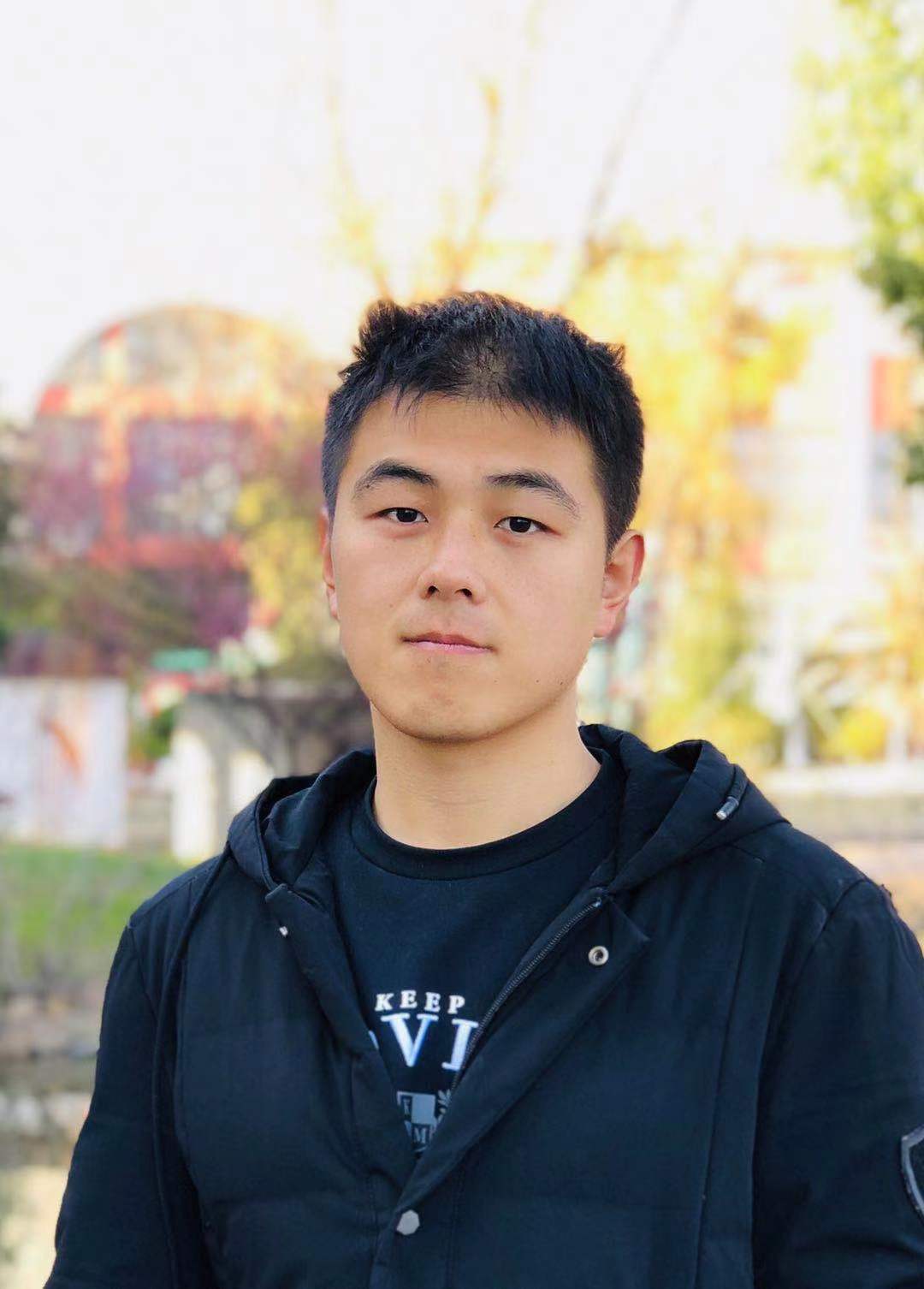}}]{Liang Xie}
Liang Xie is currently pursuing his PhD degree in the College of Computer Science and Technology at Zhejiang University and his research interests include computer vision and machine learning.
\end{IEEEbiography}

\begin{IEEEbiography}[{\includegraphics[width=1in,height=1.25in,clip,keepaspectratio]{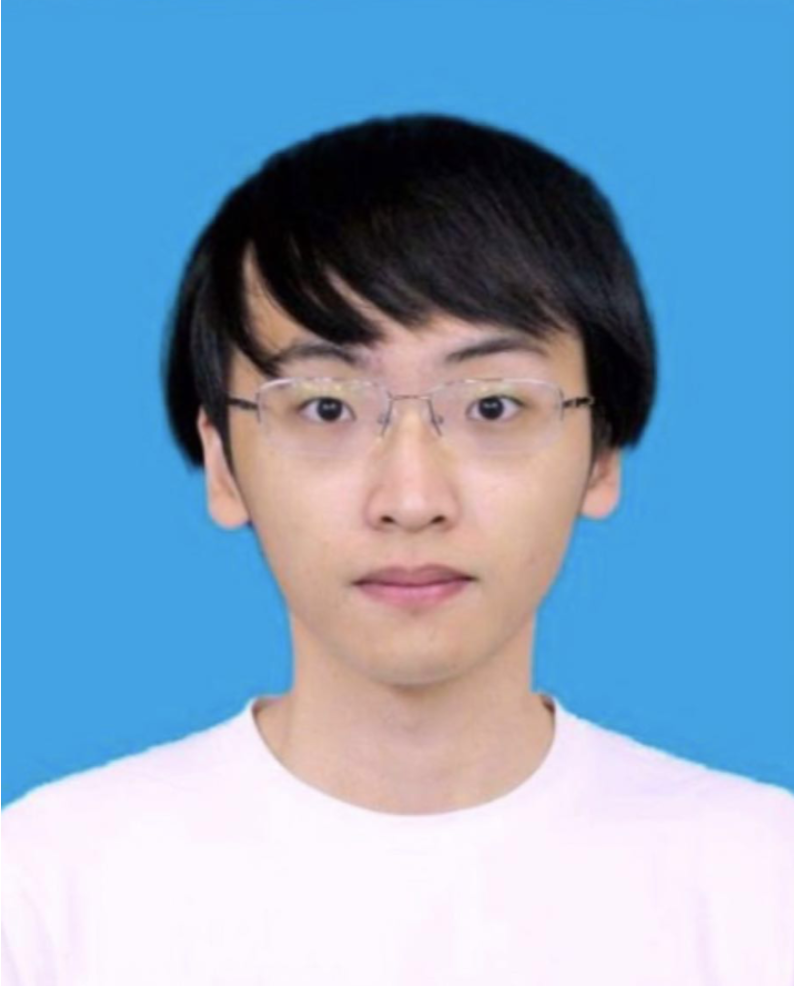}}]{Guodong~Xu}
  Guodong Xu is currently pursuing the M.S. degree in the College of Computer Science and Technology at Zhejiang University. His research interests include computer vision, deep learning and 3D scene understanding.
\end{IEEEbiography}

\begin{IEEEbiography}[{\includegraphics[width=1in,height=1.25in,clip,keepaspectratio]{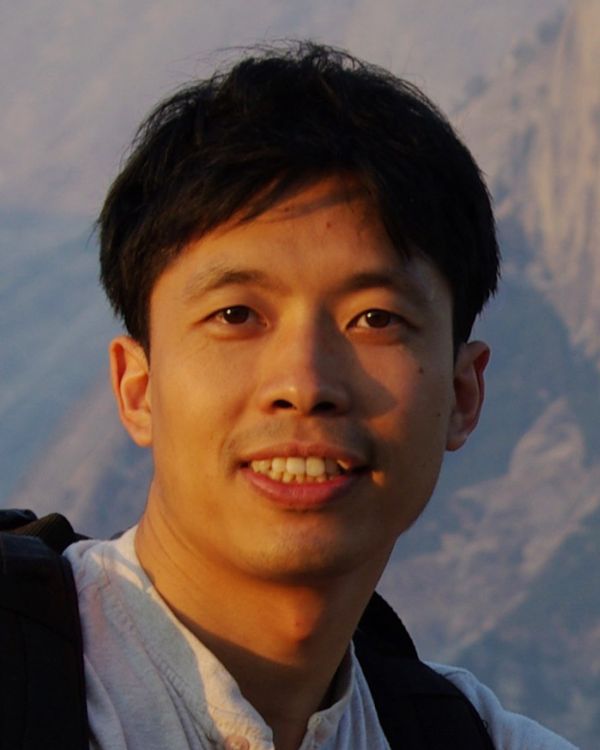}}]{Deng~Cai}
  Deng Cai He is currently a full professor in the College of Computer Science at Zhejiang University, China. He received the PhD degree from University of Illinois at Urbana Champaign. His research inter- ests include machine learning, computer vision, data mining and information retrieval.
\end{IEEEbiography}

\begin{IEEEbiography}[{\includegraphics[width=1in,height=1.25in,clip,keepaspectratio]{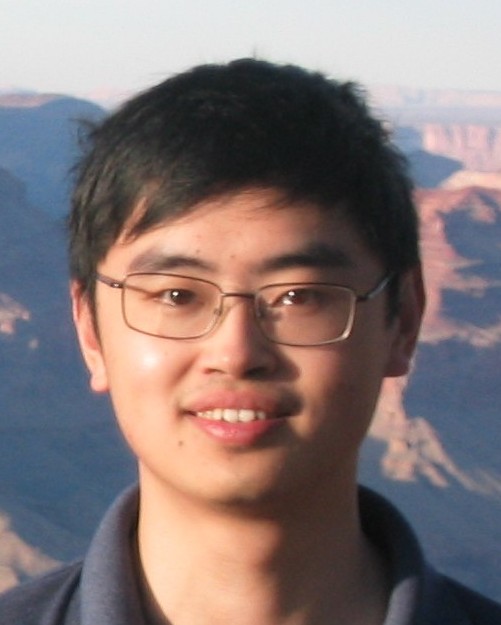}}]{Xiaofei~He}
  Xiaofei He He is currently a full professor in the College of Computer Science at Zhejiang University, China. He received the BS degree from Zhejiang University, China, in 2000 and the PhD degree from University of Chicago in 2005, both in computer science. After my PhD, He joined Yahoo Research Labs as a research scientist. He joined Zhejiang University in 2007.
\end{IEEEbiography}




\end{document}